%% file: paper.tex
\documentclass[]{bytedance_seed}

\usepackage[toc,page,header]{appendix}
\usepackage{minitoc}
\usepackage{amsmath}

\titleformat*{\paragraph}{\bfseries}

\title{Unfolding Scientific Papers into Multi-Turn Generation Trajectories for Continued Pre-Training}

\author[1,2,\ddagger,\dagger]{Qiankai Xu}
\author[1,3,\ddagger]{Qiguang Chen}
\author[1]{Zixin Su}
\author[1]{Wenhao Huang}
\author[1]{Yue Gao}
\author[2,4]{\newline Jiaheng Liu}
\author[1,\dagger]{Ge Zhang}

\affiliation[1]{ByteDance Seed}
\affiliation[2]{Nanjing University}
\affiliation[3]{Evolvent AI}
\affiliation[4]{TokenWave.AI}

\contribution[\ddagger]{Work done at ByteDance}
\contribution[\dagger]{Corresponding authors}

\abstract{
A recent line of synthetic-data work reconstructs the thinking behind existing text rather than rewriting the text itself, but it operates on short web passages, recovers only local thoughts, and leaves the structure of whole documents untouched. Scientific papers are written to a clear and largely uniform structure and make a natural substrate for lifting this paradigm to the document level. We present a pipeline that unfolds each paper into a multi-turn generation trajectory in which a teacher model reconstructs the writing process of the whole paper: a writing request, a global plan, and pre-writing deliberation for each section. All section texts and the abstract are kept verbatim from the source paper. We apply the pipeline to 1.8M quality-filtered arXiv papers and obtain a 60B-token corpus for continued pre-training (CPT) that is roughly twice the source text. The same reverse construction extends to instruction data and evaluation. We build an SFT dataset of 200K samples using answers derived from paper text. We also use held-out papers to construct PAW-Bench, a benchmark of 2,940 academic writing tasks with per-task rubrics and checklists. In controlled experiments, CPT on our corpus followed by SFT on public datasets improves writing performance while preserving general reasoning and improving long document reading. Replacing part of the writing SFT data with our synthetic instruction data further improves performance on PAW-Bench.
}

\date{August 2, 2026}

\correspondence{Qiankai Xu at \email{qiankaixu6@gmail.com}, Ge Zhang at \email{zhangge.eli@bytedance.com}}

\begin{document}
\maketitle

\input{sections/introduction}
\input{sections/relatedwork}
\input{sections/method}
\input{sections/experiments}
\input{sections/conclusion}

\clearpage

\bibliographystyle{plainnat}
\bibliography{main}

\clearpage

\beginappendix

\input{sections/appendix}

\end{document}

%% file: sections/introduction.tex
\section{Introduction}
\label{sec:intro}

Synthetic data plays an increasing role in training large language models as demand for human text grows. Pretraining and mid-training consume tens of billions of tokens. Early approaches rewrite web text into cleaner prose, textbook passages, or question and answer pairs, then train on these alongside the original text \citep{maini2024wrap,yang2024entigraph}.

Recent work keeps the original text and reconstructs the \textbf{thinking behind it}. Although finished documents often give little account of the decisions that shaped them, a model can use their results and arguments to infer plausible intermediate thoughts. Training on these thoughts alongside the original improves data efficiency and reasoning \citep{ruan2025latent,ishibashi2025mining,wang2025thinking,kim2026megadocs}.

\begin{samepage}
Existing methods mostly use short web documents, including mathematical, educational, and general text of one or two thousand tokens \citep{ruan2025latent,ishibashi2025mining,wang2025thinking,kim2026megadocs}. Their construction usually follows two patterns:
\begin{enumerate}
    \item \textbf{Fixed text segments.} Reasoning is inserted around blocks with a fixed number of sentences or segments of equal length.
    \item \textbf{Local reconstruction.} Synthesized thoughts are attached before or after a passage. The training unit is the passage and its associated thinking.
\end{enumerate}
\end{samepage}
These patterns suit short passages, while planning and drafting a complete scientific paper require a representation that connects its sections. We study this setting and evaluate its effect on writing as well as reasoning.

\begin{figure}[!t]
\centering
\includegraphics[width=\textwidth]{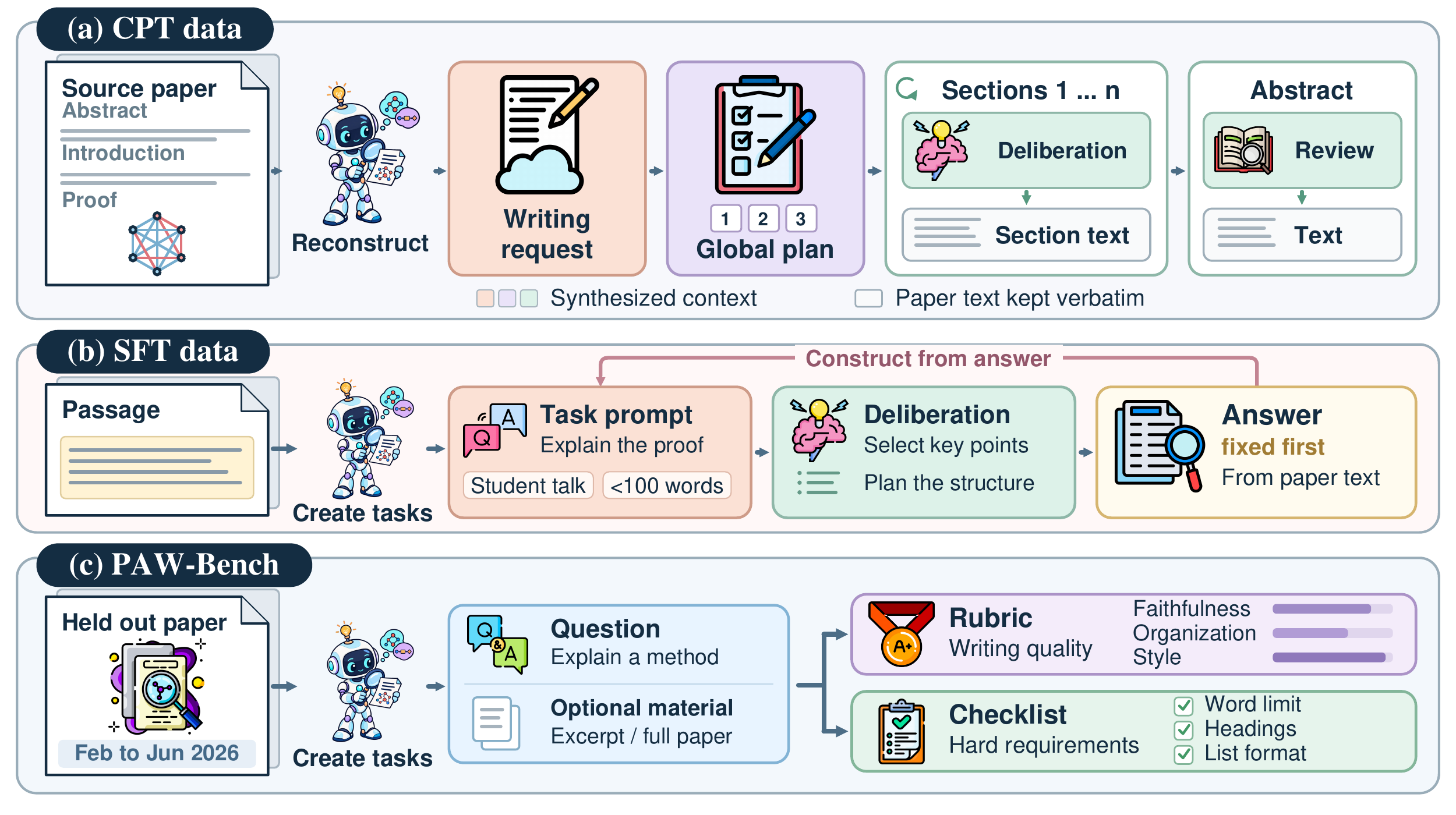}
\caption{From scientific papers to training and evaluation data. \textbf{Top:} CPT reconstructs a writing request, a global plan, and deliberation before each section. The sections and abstract are kept verbatim, with the abstract placed last. \textbf{Middle:} SFT first derives an answer from paper text, then constructs the task prompt and deliberation. The return arrow shows this construction order; forward arrows show the final training sample. \textbf{Bottom:} Papers reserved for evaluation yield a question with optional source material and two sets of criteria: a rubric for writing quality and a checklist for hard requirements. Task details and scores are illustrative.}
\label{fig:trajectory}
\end{figure}

A paper's introduction, methods, experiments, and conclusion have distinct roles, making section boundaries useful for reconstructing a writing process. The document itself supplies the scientific content needed to infer a plausible plan and explain what each section should establish.

Our pipeline \textbf{unfolds} each paper into a multi-turn generation trajectory (\Cref{fig:trajectory}(a)). Beginning with a writing request and a global plan with section outlines, the trajectory places deliberation before each section and a review of the completed paper before the abstract. An LLM generates the request, plan, and deliberations; the sections and abstract are copied verbatim. Applied to 1.8M arXiv papers with three generators of different sizes, the pipeline turns 30B tokens of paper text into 57--60B tokens of trajectories per generator. The median training document grows from 11K to 28--29K tokens.

The same construction produces instruction data and evaluation tasks from real paper content. For SFT, we derive an answer from a passage and construct its task and thinking (\Cref{fig:trajectory}(b)). For PAW-Bench, papers reserved for evaluation supply tasks with rubrics and checklists for grading answers (\Cref{fig:trajectory}(c)). One arXiv source thus supplies a CPT corpus, an SFT dataset, and an academic writing benchmark.

Our experiments compare CPT on trajectories with CPT on the same papers as plain text, followed by identical SFT recipes. The trajectories improve writing while preserving general reasoning and extending the range of document lengths seen during training. Adding our instruction data at the SFT stage brings further writing gains (\Cref{sec:main}).

Our contributions are:
\begin{enumerate}
    \item \textbf{A scalable pipeline} that unfolds scientific papers into multi-turn generation trajectories with a global plan and deliberation before each section. Processing 1.8M papers turns 30B tokens of paper text into a CPT corpus of 60B tokens.
    \item \textbf{Instruction data and evaluation tasks} built with the same construction (\Cref{fig:trajectory}(b) and (c)). Answers derived from paper text yield an SFT dataset of 200K samples. Papers reserved for evaluation yield PAW-Bench (Paper-Anchored Writing Benchmark), with 2,940 academic writing tasks and a rubric and checklist for each.
    \item \textbf{Controlled experiments} using the same papers as plain text to measure the effect of unfolding. The trajectories improve writing under both SFT recipes, preserve general reasoning, and benefit paper comprehension and long context performance.
\end{enumerate}

%% file: sections/relatedwork.tex
\section{Related Work}
\label{sec:related}

\paragraph{Synthetic data for model training.}
Synthetic data is used throughout model training. Pretraining methods rewrite web documents or expand text around entity relations \citep{maini2024wrap,yang2024entigraph}, while posttraining methods generate instructions \citep{wang2023selfinstruct,ge2024personahub}. Repeated training on generated data can reduce coverage of less frequent content. Training on edited real text is more robust \citep{shumailov2024collapse,zhu2025toedit}.

\paragraph{Reconstructing training signal from real text.}
Several methods infer the thinking behind web text and combine it with the original for training, at scales up to 100B tokens \citep{ruan2025latent,ishibashi2025mining,wang2025thinking}. Megadocs inserts reasoning at evenly spaced points to form longer training documents \citep{kim2026megadocs}. For open-ended writing, REER reconstructs deliberation that explains a known good answer \citep{wang2025reer}. Other methods take real text as the answer and derive a request that would elicit it \citep{li2024backtranslation}. This approach has been used at pretraining scale \citep{patel2026fineinstructions} and extended to writing through outlines derived from full articles \citep{liang2024planning}. Related work reconstructs development trajectories for entire GitHub projects \citep{zeng2026reconstruction}. Work on natural language text largely uses short web documents and local annotations. We use the section structure of scientific papers to represent planning and drafting as multi-turn trajectories.

\paragraph{Scientific literature: adaptation and evaluation.}
Models adapted to scientific text are trained on cleaned prose from papers and textbooks \citep{li2025scilitllm}. Our data adds a reconstructed writing process around this prose. Existing writing benchmarks generate criteria for each prompt \citep{wu2025writingbench} or build tasks from arXiv at several levels of abstraction \citep{zhang2025academiceval}. PAW-Bench derives each task from a specific paper and reports both writing quality and compliance with a checklist, whose items are assessed by code or an LLM.

%% file: sections/method.tex
\section{Method}
\label{sec:method}

We construct training data and evaluation tasks from existing paper text. For the CPT corpus, a model reconstructs a writing process around a complete paper (\Cref{sec:cpt}). For the SFT data, it derives a task and deliberation from a passage (\Cref{sec:sft}). For PAW-Bench, it derives a task and grading criteria from a paper reserved for evaluation (\Cref{sec:paw}). All three use arXiv papers and share the preprocessing pipeline described in \Cref{sec:source}.

\begin{figure}[!ht]
\centering
\includegraphics[width=0.90\textwidth]{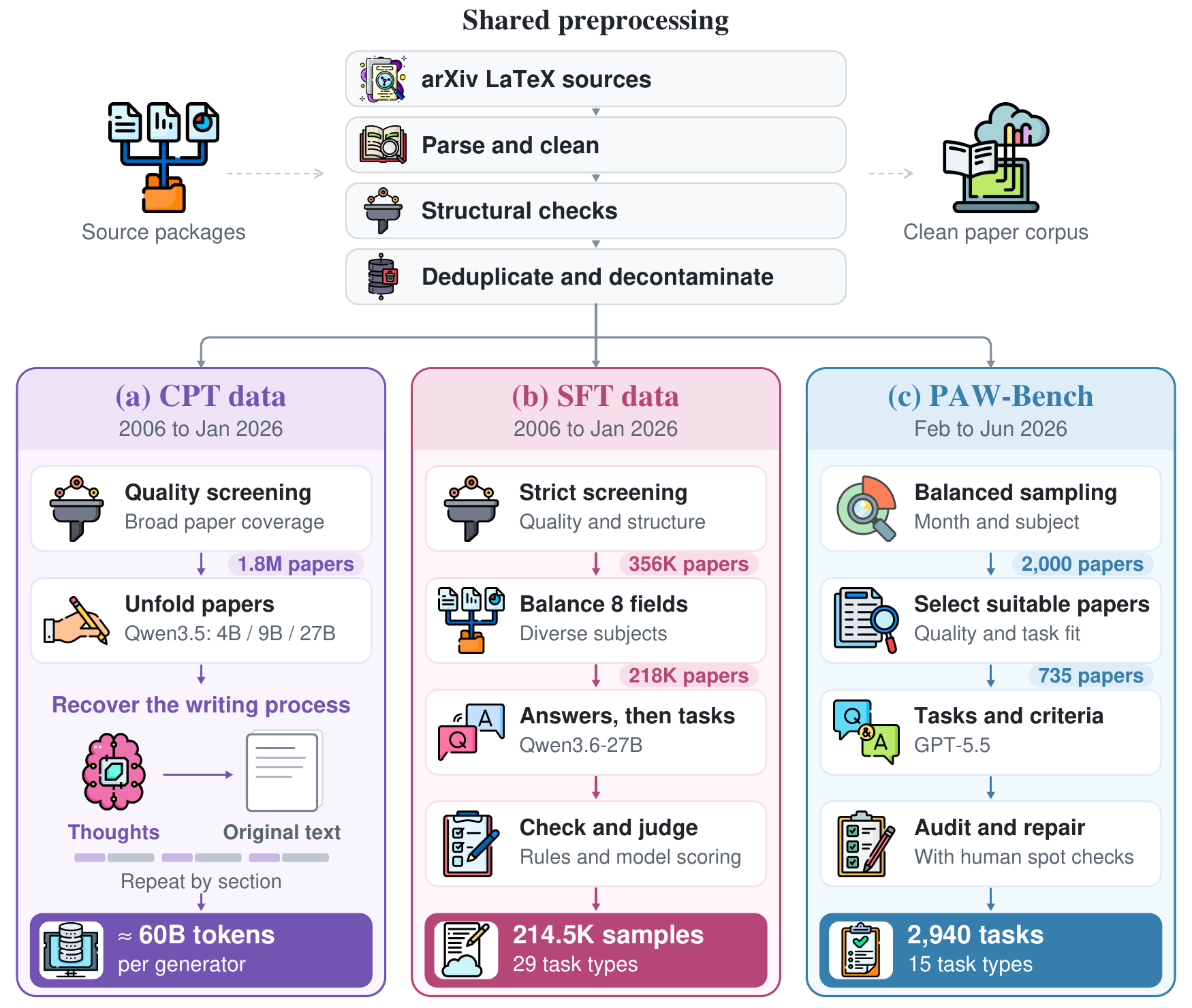}
\caption{The data pipeline. \textbf{Top:} Shared preprocessing cleans arXiv sources and removes malformed papers, duplicates, and benchmark overlaps. \textbf{Bottom~left:} CPT uses 1.8M papers to reconstruct writing trajectories. \textbf{Bottom~middle:} SFT applies stricter screening and subject balancing before task construction and quality checks. \textbf{Bottom~right:} PAW-Bench uses GPT-5.5 to generate tasks and grading criteria from papers posted after the training window. Counts show the pool size after each stage.}
\label{fig:pipeline}
\end{figure}

\subsection{Preparing the Paper Source}
\label{sec:source}

\paragraph{Shared preprocessing.} The CPT data, SFT data, and PAW-Bench share a pipeline for preprocessing and filtering public arXiv papers (\Cref{fig:pipeline}). For each paper, we locate the main source file, merge included files, and expand custom macros. We retain the body and remove references, acknowledgments, appendices, and funding statements. We also remove citation markers, figure environments, and layout commands, while retaining equations and tables. This reduces references to material that the model would not receive during training. Structural rules exclude errata, supplements, and papers with an empty body, a length outside the allowed range, or too few sections. We deduplicate by body text, remove papers containing complete questions from downstream benchmarks, and retain those with high LLM quality scores.

\paragraph{Training sources.} The CPT and SFT sources cover arXiv papers from 2006 through January 2026 and use different filtering thresholds. For CPT, a small model removes clearly defective papers, leaving 1.8M. SFT applies stricter limits on body length, language, section count, letter ratio, LaTeX command density, and the proportion of repeated paragraphs. It also requires an opening section that serves as an introduction and excludes papers with residual citations or figure commands after cleaning. A stronger model scores the remaining papers, leaving 356K. To balance arXiv's eight subject groups, we retain the smaller groups in full and select the papers with the highest scores from larger groups, mainly computer science, physics, and mathematics. The final pool contains 220K papers.

\paragraph{Evaluation source.} PAW-Bench uses papers from February through June 2026, after the training source window. Its papers are therefore separate from those used to construct the training data. A strong model rates 2,000 papers, balanced by month and category, for quality and suitability as task material. Retaining the top half of each subject group leaves 735 papers for task construction.

\subsection{CPT Data: Unfolding Papers into Multi-Turn Trajectories}
\label{sec:cpt}

\paragraph{Trajectory definition.} Let $P=(t,s_1,\ldots,s_n,a)$ denote a cleaned paper with title $t$, retained sections $s_i$, and abstract $a$. We split the body at section boundaries and discard sections that are too short. Unfolding constructs a writing request $q$, a planning block $p$ with section outlines, a deliberation $c_i$ before each section, and a review $c_a$ before the abstract. The resulting sequence is
\begin{equation}
\tau(P)=q\oplus p\oplus c_1\oplus s_1\oplus\cdots\oplus c_n\oplus s_n\oplus c_a\oplus a,
\label{eq:trajectory}
\end{equation}
where $\oplus$ denotes ordered text concatenation with the appropriate separators. The sections $s_i$ and abstract $a$ are copied verbatim from the cleaned paper; the surrounding context is synthesized (\Cref{fig:trajectory}(a)). The sequence follows the order of planning, drafting the sections, and writing the abstract last.

\paragraph{Reconstructing the writing context.} Let $G_\phi$ be the generator, with superscripts denoting the prompt used at each stage. It first summarizes each section as $u_i$, collecting the summaries in $U=(u_1,\ldots,u_n)$. The main dependencies are
\begin{equation}
\begin{aligned}
u_i &\sim G_\phi^{\mathrm{sum}}(\cdot\mid s_i),
& q &\sim G_\phi^{\mathrm{req}}(\cdot\mid t,U),\\
g &\sim G_\phi^{\mathrm{plan}}(\cdot\mid t,q,U),
& p &=g\oplus u_1\oplus\cdots\oplus u_n,\\
c_i &\sim G_\phi^{\mathrm{sec}}(\cdot\mid t,s_i,U_{<i},U_{>i}),
& c_a &\sim G_\phi^{\mathrm{abs}}(\cdot\mid t,a,U),
\end{aligned}
\label{eq:reconstruction}
\end{equation}
where $U_{<i}$ and $U_{>i}$ contain summaries of the preceding and following sections. For simplicity, the notation omits section labels, prompt instructions, and input length limits. The global plan $g$ describes the paper's goals, evidence, and organization; the summaries serve as section outlines in $p$. Each $c_i$ uses the current section and summaries of the surrounding sections to explain what the section should establish and how it contributes to the argument. The final review $c_a$ identifies the findings to include in the abstract. Using summaries reduces the cost of processing millions of papers. The prompts allow varied formats, lengths, and perspectives to reduce repetitive synthetic text \citep{niklaus2026finephrase}.

\paragraph{Training on the unfolded sequence.} Let $Y$ be the assistant target retained after truncating the complete sequence in \Cref{eq:trajectory}. With $q$ and $Y$ tokenized, the trajectory contributes the autoregressive loss for a student model $p_\theta$:
\begin{equation}
\ell_{\mathrm{traj}}(\theta;q,Y)
=-\sum_{j=1}^{|Y|}\log p_\theta(Y_j\mid q,Y_{<j}).
\label{eq:trajectory-loss}
\end{equation}
The request $q$ provides context and is masked from the loss, while the planning, deliberation, and retained paper text all receive supervision. Although the generator has access to the completed paper during reconstruction, the student predicts each token from the preceding context during training. CPT mixes these trajectories with general text under the shared training budget in \Cref{sec:setup}.

\begin{figure}[!ht]
\centering
\includegraphics[width=0.58\textwidth]{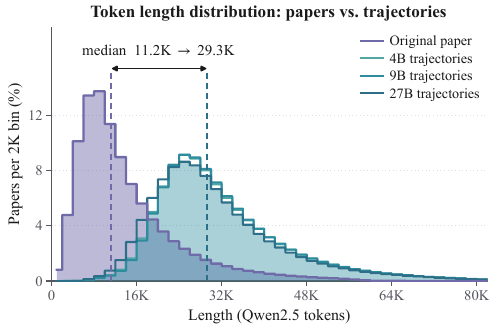}
\caption{Token length distributions of source papers and trajectories from all three generators (1.8M papers per generator). Dashed lines mark the paper median (11.2K) and the 4B trajectory median (29.3K). The trajectory medians span 28--29K tokens, about an order of magnitude longer than the 1--2K web documents used in prior reconstruction work. The horizontal axis ends at 84K; 2.6--2.9\% of trajectories are longer, up to the 128K generation limit.}
\label{fig:length}
\end{figure}

\paragraph{Generation.} We use Qwen3.5-4B, 9B, and 27B \citep{qwen2026qwen35} to process the 1.8M papers, sampling at temperature 1.0 with a budget of 8,192 tokens per call. \Cref{app:prompts,app:cpt-pipeline} give the full prompts and decoding settings. The source bodies total about 30B tokens, and unfolding yields 57--60B tokens per generator, roughly a $2\times$ expansion. The median body length is 11.2K tokens; the corresponding trajectory medians are 29.3K, 29.4K, and 28.3K (\Cref{fig:length}). Prior work on reconstructing web text uses documents of one or two thousand tokens. Unfolding papers thus provides substantially longer training examples, with planning and deliberation throughout each document.

\begin{figure}[!ht]
\centering
\includegraphics[width=\textwidth]{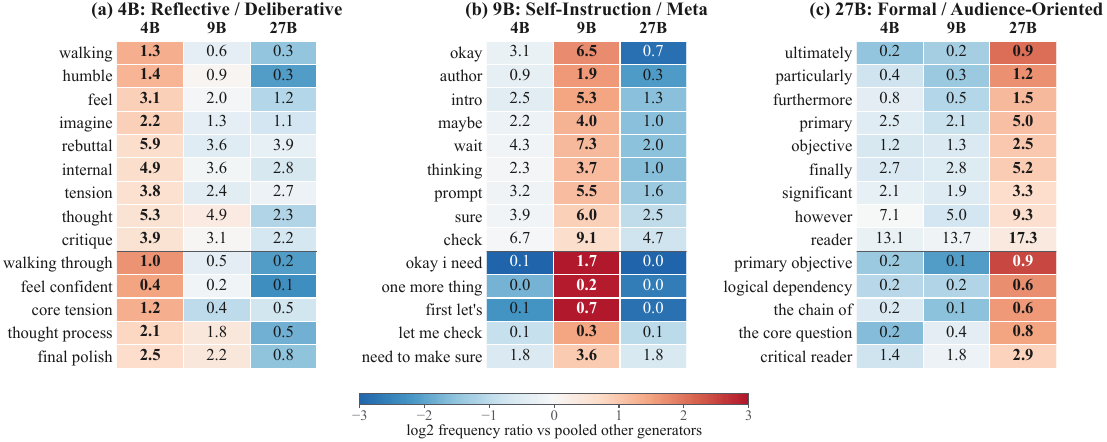}
\caption{Lexical patterns in trajectories generated from the same 10,000 papers by Qwen3.5-4B, 9B, and 27B, counting only synthesized text. \textbf{Left:} Reflective and deliberative expressions characteristic of 4B. \textbf{Middle:} Expressions used by 9B to guide and check its own thinking. \textbf{Right:} Formal expressions and references to the reader characteristic of 27B. Cell values are exact occurrences per 10K reasoning words; colors show the log$_2$ frequency ratio relative to the other two generators pooled together. The final five rows in each panel are contiguous phrases; the others are single words.}
\label{fig:lexical}
\end{figure}

\paragraph{Generator writing styles.} The median 27B trajectory is about 1K tokens shorter than the 4B and 9B trajectories. Word and phrase counts over 10,000 papers, totaling about 315M reasoning words, also show differences in style (\Cref{fig:lexical}). The 4B model narrates its thinking (\textit{imagine}, \textit{walking through}), 9B uses reminders and checks (\textit{okay}, \textit{wait}, \textit{let me check}), and 27B uses more formal expressions (\textit{however}, \textit{ultimately}, \textit{critical reader}). The shorter 27B trajectories contain less narration of the reasoning process. All three generators follow the same trajectory structure.

\subsection{SFT Data: Deriving Tasks Backwards from Paper Text}
\label{sec:sft}

\paragraph{Sample structure.} Each SFT sample is a QA pair with one user turn and one assistant turn. The user poses a writing or reading task, optionally providing a full paper or excerpt, and the assistant produces deliberation followed by an answer. We design 29 task types covering academic writing and reading. They form three families according to how the answer is derived: rewriting tasks adapt a passage into an output such as a reproduction checklist; QA tasks derive a question and answer from the material; and reorganizing tasks combine several passages into a new form, such as a decision memo.

\begin{figure}[!ht]
\centering
\includegraphics[width=\textwidth]{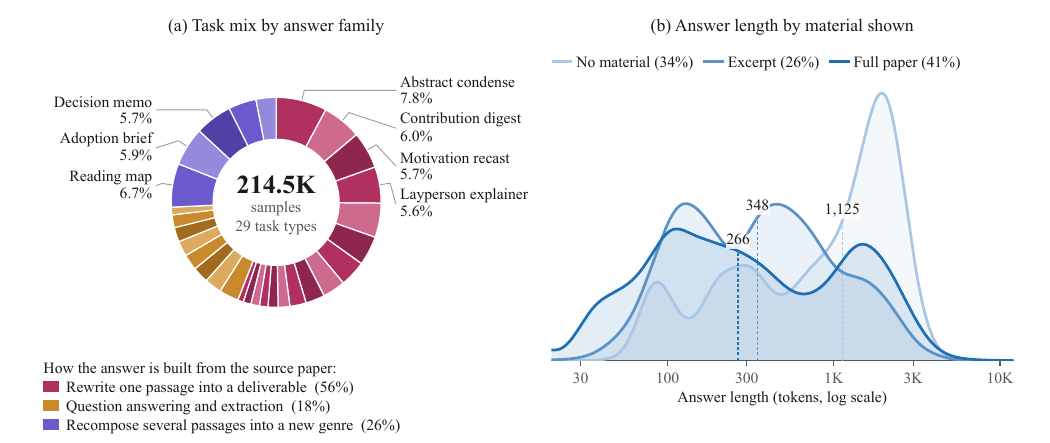}
\caption{Composition of the SFT dataset (214.5K samples from 148.6K papers). \textbf{Left:} The 29 task types, grouped by how their answers are derived from the paper. Types above 5\% are labeled. \textbf{Right:} Answer length distributions by the material provided to the model; dashed lines mark medians. Tasks without paper material have the longest answers. Tasks that provide the full paper tend to ask for compact extraction and summary outputs.}
\label{fig:sftmix}
\end{figure}

\paragraph{Composition.} \Cref{fig:sftmix} shows the distribution of task types. The largest type accounts for 7.8\% of the data, and the three families contribute 56\%, 18\%, and 26\%, respectively. Rewriting tasks provide the full paper in half of the samples, QA tasks always provide paper material, and five of every six reorganizing samples provide none. Across the dataset, 34\% of samples require writing from the question alone and have the longest answers, with a median of about 1.1K tokens (\Cref{fig:sftmix}(b)). Tasks with paper material mostly ask for concise extraction or summaries, with median answer lengths near 300 tokens.

\paragraph{Constructing tasks from answers.} For rewriting and reorganizing tasks, we first turn selected paper content into a standalone answer $y$, then derive its request $q$ and deliberation $r$ (\Cref{fig:trajectory}(b)). Let $G_\psi$ denote the SFT generator, $\omega$ the task type, audience, material settings, and format requirements, and $e$ the source excerpt provided when constructing the deliberation. The synthesis order is
\begin{equation}
\begin{aligned}
y &\sim G_\psi^{\mathrm{ans}}(\cdot\mid P,\omega),\\
q &\sim G_\psi^{\mathrm{req}}(\cdot\mid y,\omega),\\
r &\sim G_\psi^{\mathrm{delib}}(\cdot\mid q,y,e).
\end{aligned}
\label{eq:sft-construction}
\end{equation}
The request's length and format constraints are derived from the completed answer and checked against it. For QA tasks, the first two stages jointly generate $(q,y)$ from the supplied material. Although the deliberation generator sees the answer during construction, its instructions require a plan based on the user's task and available material, without referring to the supplied answer.

\paragraph{Training samples and filtering.} If $M$ denotes the material shown to the student, the final sample is $((q,M),\,r\oplus y)$, with $M=\emptyset$ when no paper text is attached. The construction excerpt $e$ is also empty for these tasks. SFT supervises the assistant's deliberation and answer and masks the user input, following the conditional prediction objective in \Cref{eq:trajectory-loss}. We sample three suitable task types for each paper. Programmatic checks remove failures such as residual LaTeX or deliberation that mentions ``the provided answer''. An LLM scores the remaining samples, and we retain those with higher scores. Using Qwen3.6-27B \citep{qwen2026qwen36} as the generator yields about 200K samples. In \Cref{sec:main}, we combine 10K samples focused on writing from this dataset with other SFT data.

\subsection{PAW-Bench: Paper-Anchored Writing Benchmark}
\label{sec:paw}

\paragraph{Task construction.} GPT-5.5 reads a paper reserved for evaluation and generates a question and grading criteria in the same pass. The paper supplies content for the question and evidence for assessing the answer's completeness.

\paragraph{Task design.} We generate four tasks for each of the 735 papers selected in \Cref{sec:source}, giving 2,940 tasks across 15 types related to the SFT types. Nine types provide the full paper or selected excerpts and test understanding and writing based on the material. Six provide no paper material and test general academic writing in realistic work scenarios. We assign task types before generation to balance their counts. \Cref{tab:pawtypes} lists all fifteen types and the material they provide. Half of the tasks provide no material, about a quarter provide an excerpt, and about a quarter provide the full paper (1,470, 764, and 706 tasks). The subject distribution is approximately uniform, ranging from 15\% for computer science to 9\% for quantitative biology. For each task, we randomly select the input material format, required response length, and formatting requirements from the options defined for its task type.

\begin{table}[!ht]
\centering
\footnotesize
\setlength{\tabcolsep}{6pt}
\begin{tabular}{@{}llc@{}}
\toprule
Task type & Objective & Count \\
\midrule
\multicolumn{3}{@{}l}{\emph{Without paper material: the query describes the scenario}} \\[1.5pt]
\texttt{academic\_abstract\_without\_source} & Write an academic abstract from a scenario. & 244 \\
\texttt{research\_proposal\_framework} & Construct a research proposal framework. & 245 \\
\texttt{technical\_report\_without\_source} & Produce a technical report for a stated problem. & 246 \\
\texttt{experiment\_discussion\_without\_source} & Discuss an experimental setup and findings. & 244 \\
\texttt{literature\_review\_outline\_without\_source} & Design a literature review outline. & 245 \\
\texttt{policy\_or\_industry\_analysis\_without\_source} & Write policy or industry analysis using evidence. & 246 \\
\midrule
\multicolumn{3}{@{}l}{\emph{With paper material: the query includes an excerpt or the full paper}} \\[1.5pt]
\texttt{abstract\_from\_paper} & Write an abstract. & 163 \\
\texttt{tldr\_contribution\_bullets} & Produce concise contribution bullets. & 164 \\
\texttt{introduction\_rewrite} & Rewrite an introduction for a specified goal. & 163 \\
\texttt{method\_explanation} & Explain the method to a target audience. & 162 \\
\texttt{experiment\_summary} & Summarize experimental design and evidence. & 163 \\
\texttt{limitation\_future\_work} & Synthesize limitations and future work. & 164 \\
\texttt{review\_critique} & Write a structured review and critique. & 164 \\
\texttt{technical\_report\_from\_paper} & Produce a technical report. & 164 \\
\texttt{plain\_language\_summary} & Explain the work to non-specialists. & 163 \\
\bottomrule
\end{tabular}
\caption{PAW-Bench task composition. Requested answer lengths span 20--2,100 words with median 1,000.}
\label{tab:pawtypes}
\end{table}

\begin{figure}[!ht]
\centering
\includegraphics[width=\textwidth]{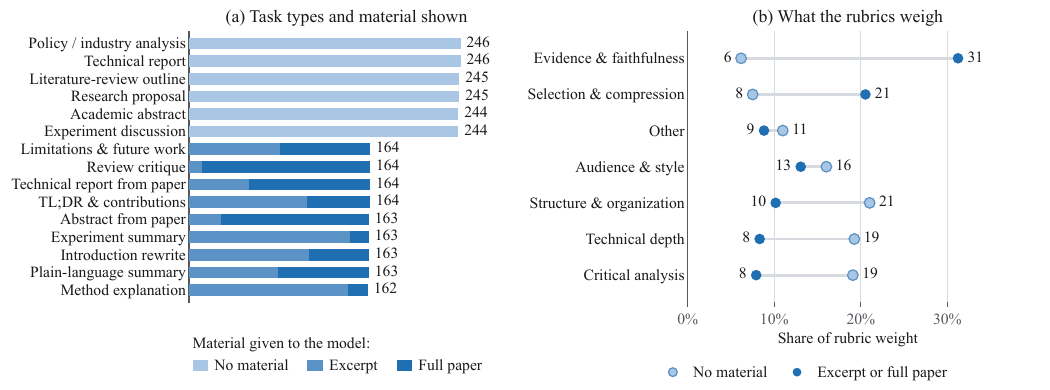}
\caption{Composition of PAW-Bench (2,940 tasks, four per paper from 735 papers reserved for evaluation). \textbf{Left:} Counts for the 15 task types, with color indicating whether the model receives no paper material, an excerpt, or the full paper. \textbf{Right:} Distribution of rubric weight across assessment themes, comparing tasks with and without paper material. The 4,898 distinct dimension names across all tasks are grouped by keyword; the remaining names fall into Other.}
\label{fig:pawmix}
\end{figure}

\paragraph{Grading criteria.} Weighted rubric dimensions assess qualities such as content coverage, organization, and faithfulness to the material. A separate checklist specifies required content and constraints on length and format. About half of these requirements can be checked by code. For a response $z$, let $s_k(z)\in\{1,\ldots,10\}$ be the judge's score on rubric dimension $k$, and let $b_j(z)\in\{0,1\}$ indicate whether checklist item $j$ passes. With nonnegative weights normalized separately so that $\sum_k w_k=\sum_j v_j=1$, we report
\begin{equation}
\operatorname{Rubric}(z)=10\sum_{k=1}^{K}w_k s_k(z),
\qquad
\operatorname{Checklist}(z)=\sum_{j=1}^{J}v_j b_j(z).
\label{eq:paw-scores}
\end{equation}
Each $b_j$ is determined by code or an LLM, depending on the requirement. Benchmark scores average each metric over tasks. Rubric dimensions and weights vary by task and by the material supplied (\Cref{fig:pawmix}(b)). Tasks with paper material assign 31\% of their rubric weight to faithfulness to the evidence, compared with 6\% for tasks without it. The latter emphasize structure, technical substance, and critical reasoning. We checked the full task pool and repaired identified issues over several rounds, then inspected a sample manually. PAW-Bench is one of the four writing benchmarks in \Cref{sec:main} and derives tasks from paper content using the same approach as our training data.

%% file: sections/experiments.tex
\FloatBarrier
\section{Experiments}
\label{sec:exp}

\subsection{Setup}
\label{sec:setup}

\paragraph{Training recipe.} All experiments start from Qwen2.5-7B \citep{qwen2024qwen25} and follow three stages: CPT, SFT, and evaluation. We use a learning rate of 2e-5 with cosine decay to 10\% of the peak, one epoch, sequences truncated at 64K, about 4M tokens per step, weight decay 0.01, and gradient clipping 1.0. We choose a small learning rate to limit changes to the base model during adaptation to the narrower CPT distribution over 50B tokens. Every run uses seed 42.

\paragraph{Mixtures and controls.} From each generator's output, we sample 30B tokens and mix them with 20B tokens of FineWeb-Edu \citep{penedo2024fineweb}, giving three training sets of 50B tokens each. We include general text to reduce forgetting during continued pretraining on academic data \citep{ibrahim2024continual}. Two controls use the same budget: 50B tokens of FineWeb-Edu, and 30B tokens of raw paper text (cleaned as in \Cref{sec:source}) plus 20B of FineWeb-Edu. Comparing the latter with the trajectory mixtures tests the contribution of our construction while keeping the source papers fixed. All five runs share the same hyperparameters.

\paragraph{Naming.} Our-Data-4B/9B/27B CPT denotes the models trained on data from the three generators; FineWeb-Edu CPT and Plain-Paper CPT denote the controls. \Cref{sec:main,sec:reasoning,sec:longctx} each pair these checkpoints with an open SFT dataset for a different ability. We hold the SFT recipe fixed within each comparison to measure the contribution of the CPT corpus.

\begin{table}[!ht]
\centering
\footnotesize
\setlength{\tabcolsep}{3pt}
\begin{tabular}{@{}lcccccccc@{}}
\toprule
\multirow{2}{*}{Model} & \multicolumn{2}{c}{WritingBench} & \multicolumn{2}{c}{PAW-Bench} & \multicolumn{2}{c}{HelloBench} & \multicolumn{1}{c}{LongBench-} & \multicolumn{1}{c}{\multirow{2}{*}{Avg.}} \\
\cmidrule(lr){2-3}\cmidrule(lr){4-5}\cmidrule(lr){6-7}
& \multicolumn{1}{c}{Full} & \multicolumn{1}{c}{Acad.\,\&\,Eng.} & \multicolumn{1}{c}{Rubric} & \multicolumn{1}{c}{Checklist} & \multicolumn{1}{c}{Open QA} & \multicolumn{1}{c}{Heuristic} & \multicolumn{1}{c}{Write} & \\
\midrule
Qwen2.5-7B-Instruct & 47.3 & 49.8 & 59.09 & 0.620 & 42.66 & 43.93 & 54.10 & 50.95 \\
LongWriter-8B & 47.0 & 48.4 & 56.54 & 0.524 & 40.66 & 45.83 & 57.74 & 51.13 \\
DeepWriting SFT & 49.3 & 50.0 & 55.42 & 0.529 & 40.82 & 43.40 & \textbf{60.76} & 51.90 \\
FineWeb-Edu CPT + DeepWriting SFT & 47.7 & 48.2 & 53.71 & 0.535 & 41.49 & 43.56 & 60.35 & 51.08 \\
Plain-Paper CPT + DeepWriting SFT & 49.2 & 50.1 & 57.51 & 0.540 & 41.85 & 42.49 & 59.65 & 52.12 \\
\textbf{Our-Data-4B CPT} + DeepWriting SFT & \textbf{52.5} & \textbf{53.9} & 59.47 & 0.551 & \textbf{44.13} & \textbf{48.17} & 59.27 & \underline{54.34} \\
\textbf{Our-Data-9B CPT} + DeepWriting SFT & 51.1 & \underline{53.2} & 59.34 & 0.555 & 42.40 & 43.70 & \underline{60.42} & 53.48 \\
\textbf{Our-Data-27B CPT} + DeepWriting SFT & 51.0 & 52.4 & 58.88 & 0.557 & 42.77 & \underline{46.02} & 60.07 & 53.59 \\
\textbf{Our-Mixed SFT} & 47.3 & 49.1 & \underline{63.73} & \textbf{0.650} & 42.41 & 41.37 & 56.49 & 52.35 \\
\textbf{Our-Data-27B CPT} + \textbf{Our-Mixed SFT} & \underline{51.2} & 53.1 & \textbf{64.13} & \underline{0.637} & \underline{43.45} & 46.01 & 57.47 & \textbf{54.38} \\
\bottomrule
\end{tabular}
\caption{Writing results with GPT-5.5 as the quality judge. PAW-Bench uses a fixed random sample of 300 tasks. Avg.\ is the mean of four equally weighted items: WritingBench Full, PAW-Bench rubric, HelloBench (its two subsets averaged first), and LongBench-Write. Best score per column in bold, second best underlined.}
\label{tab:writing}
\end{table}

\subsection{Main Results}
\label{sec:main}

\paragraph{Setup.} To evaluate writing after CPT (\Cref{tab:writing}), we apply SFT to the base model and each CPT model using the open dataset DeepWriting-20k \citep{wang2025reer}. We also replace 10K of its samples with writing samples from our SFT data and call the resulting mixture Our-Mixed SFT. All SFT runs follow the official DeepWriting recipe (learning rate 2e-5, constant schedule, 3 epochs). Qwen2.5-7B-Instruct and LongWriter-8B serve as references; the latter was trained specifically to generate long texts \citep{bai2024longwriter}.

\paragraph{Benchmarks.} WritingBench collects real writing requests across six domains and one hundred subdomains, including academic and engineering, finance and business, and politics and law. Prompts typically specify source material, an audience, and format requirements, and a judge scores each response against five criteria written for its prompt \citep{wu2025writingbench}. HelloBench covers five scenarios: open-ended QA, summarization, chat, text completion, and heuristic text generation. We use the two subsets closest to academic writing, open-ended QA and heuristic text generation, and score both with a manually designed checklist \citep{que2024hellobench}. LongBench-Write evaluates writing at target lengths ranging from a few hundred to over ten thousand words, with a judge scoring six dimensions that include relevance, coherence, and clarity \citep{bai2024longwriter}. PAW-Bench derives tasks from papers reserved for evaluation and scores writing quality and format compliance separately. For evaluation, we randomly sample 300 tasks without replacement from the 2,940 PAW-Bench tasks using seed 42 and use this fixed subset for all models. \Cref{sec:paw} describes the full benchmark; \Cref{app:evaluation} gives details of the evaluation subset.

\paragraph{Judges.} We use GPT-5.5 with medium reasoning effort as the quality judge for all four writing benchmarks in \Cref{tab:writing}. Rescoring the same model responses with Gemini 3.1 Pro and Seed 2.0 Pro yields the same conclusion: all five configurations using our data outperform all five reference configurations on Avg.\ under every judge (\Cref{tab:judge-robustness}). \Cref{app:additional-judges} gives the complete additional results.

\paragraph{CPT improves writing, especially on academic tasks.} In the GPT-5.5 results (\Cref{tab:writing}), CPT on 50B tokens of FineWeb-Edu scores 0.8 points below direct SFT on Avg., with lower scores on WritingBench and PAW-Bench. Plain-Paper CPT performs similarly to direct SFT (52.12 vs.\ 51.90). Training on trajectories from the same papers gives higher overall scores for all three generator sizes. The largest gains over direct SFT are on PAW-Bench (+3.5 to +4.0) and WritingBench Academic \& Engineering (+2.4 to +3.9), the two academic metrics closest to the training corpus. Full WritingBench and the two HelloBench subsets improve by smaller margins. LongBench-Write, which specifies target lengths, shows mixed results; our corpus has no explicit length targets. Our-Data-4B CPT achieves the highest Avg.\ among the CPT models with DeepWriting SFT, gaining 2.44 points over direct SFT (+4.7\%).

\paragraph{CPT and SFT provide complementary gains.} Under the same DeepWriting recipe, the three Our-Data models score 1.6--2.4 Avg.\ points above direct SFT. Replacing 10K DeepWriting samples with writing samples from our SFT data (Our-Mixed SFT) further improves academic writing: PAW-Bench rises from 55.42 to 63.73 and the checklist pass rate from 0.529 to 0.650, with lower scores on WritingBench and LongBench-Write. Adding trajectory CPT before Our-Mixed SFT raises PAW-Bench to 64.13 and gives the highest overall score, 54.38. All configurations using our CPT score at least 53.48, above the best result from SFT alone (52.35). The CPT gains therefore persist under both SFT recipes.

\paragraph{A small generator is effective.} The models trained on data from the three generators differ by 0.86 points on Avg., compared with gains of up to 2.22 over Plain-Paper CPT. Data from the smallest 4B generator gives the highest Avg.\ in this comparison. These results support using a small generator to reduce the cost of constructing the corpus.

\input{tables/judge_robustness}

\paragraph{Comparison across judges.} Plain-Paper CPT + DeepWriting SFT is the strongest reference under each judge (\Cref{tab:judge-robustness}). Averaging equally across judges, Our-Data-4B CPT + DeepWriting SFT scores 58.43 against 54.37 for this reference. Our-Data-27B CPT + Our-Mixed SFT follows at 57.72. These two configurations exchange first and second place across judges and consistently lead the references.

\begin{table}[!ht]
\centering
\small
\setlength{\tabcolsep}{5pt}
\begin{tabular}{@{}lcccccc@{}}
\toprule
\multirow{2}{*}{Model} & \multicolumn{1}{c}{MMLU} & \multicolumn{1}{c}{GPQA} & \multicolumn{1}{c}{MATH} & \multicolumn{1}{c}{AIME} & \multicolumn{1}{c}{LCB} & \multicolumn{1}{c}{\multirow{2}{*}{Avg.}} \\
& \multicolumn{1}{c}{Pro} & \multicolumn{1}{c}{-D} & \multicolumn{1}{c}{500} & \multicolumn{1}{c}{2025} & \multicolumn{1}{c}{v6} & \\
\midrule
Qwen2.5-7B-Instruct & 55.1 & 34.5 & 74.0 & 9.3 & 16.0 & 37.78 \\
OpenThoughts SFT & 61.1 & \underline{39.6} & 83.2 & \underline{26.0} & 21.7 & 46.32 \\
FineWeb-Edu CPT & 60.6 & 36.0 & 83.0 & 21.3 & 20.6 & 44.30 \\
\textbf{Our-Data-4B CPT} & \underline{62.0} & 37.6 & \underline{83.4} & \underline{26.0} & \textbf{25.1} & \textbf{46.82} \\
\textbf{Our-Data-9B CPT} & \textbf{62.2} & 38.9 & \textbf{84.4} & \textbf{26.7} & 21.1 & \underline{46.66} \\
\textbf{Our-Data-27B CPT} & 61.2 & \textbf{40.4} & 80.6 & 22.7 & \underline{23.4} & 45.66 \\
\bottomrule
\end{tabular}
\caption{General reasoning results. Every CPT model receives SFT on OpenThoughts; OpenThoughts SFT uses the same data directly on the base model. GPQA-D is GPQA-Diamond and LCB v6 is LiveCodeBench v6. Avg.\ is the mean of the five equally weighted benchmarks. AIME 2025 scores are averaged over 5 samples at temperature 0.7. Best score per column in bold, second best underlined.}
\label{tab:reasoning}
\end{table}

\subsection{Effect on Reasoning}
\label{sec:reasoning}

\paragraph{Setup.} We evaluate reasoning to measure how CPT on writing trajectories affects other abilities. Each CPT model receives SFT on the open dataset OpenThoughts \citep{guha2025openthoughts}. We then evaluate five benchmarks covering knowledge, science QA, math, and code: MMLU-Pro \citep{wang2024mmlupro}, GPQA-Diamond \citep{rein2023gpqa}, MATH500 \citep{lightman2024letsverify}, AIME 2025, and LiveCodeBench v6 \citep{jain2024livecodebench}. All SFT runs follow the official OpenThoughts training recipe (learning rate 1e-5 with cosine decay, 3 epochs).

\paragraph{Reasoning performance remains close to direct SFT.} FineWeb-Edu CPT scores below direct SFT on all five benchmarks, with Avg.\ falling from 46.32 to 44.30 (\Cref{tab:reasoning}). It also gives the lowest writing Avg.\ among the CPT runs (\Cref{tab:writing}). Replacing 30B of those tokens with our data brings reasoning Avg.\ to 45.66--46.82, close to direct SFT. This smaller decline is consistent with reduced forgetting during continued pretraining \citep{ibrahim2024continual}.

\paragraph{Generator scale has a modest effect.} Each of the three Our-Data models leads on at least one benchmark. Data from the 4B generator gives the highest Avg., at 46.82 compared with 46.32 for direct SFT.

\subsection{Paper Comprehension and Long Context}
\label{sec:longctx}

\paragraph{Setup.} We evaluate paper comprehension and long context after SFT on the no-think subset of SmolTalk2 \citep{bakouch2025smollm3}, Hugging Face's posttraining collection for SmolLM3. This subset covers general instruction following and dialogue across multiple turns, with answers that omit explicit reasoning traces. All models use the same training settings (learning rate 1e-5 with cosine decay, 2 epochs).

\paragraph{Benchmarks.} Qasper asks research questions about full NLP papers and is scored by QA F1 \citep{dasigi2021qasper}. QASA uses AI/ML papers and is scored by ROUGE-L \citep{lee2023qasa}. LongBench v2 evaluates comprehension and reasoning over long contexts, measured by accuracy \citep{bai2025longbench2}. These three evaluation sets are smaller than the writing and reasoning sets, so the results provide supporting evidence for the main findings.

\begin{table}[!ht]
\centering
\small
\setlength{\tabcolsep}{6pt}
\begin{tabular}{@{}lcccc@{}}
\toprule
\multirow{2}{*}{Model} & \multicolumn{1}{c}{\multirow{2}{*}{Qasper}} & \multicolumn{1}{c}{\multirow{2}{*}{QASA}} & \multicolumn{2}{c}{LongBench v2} \\
\cmidrule(lr){4-5}
& & & \multicolumn{1}{c}{$\leq$32k} & \multicolumn{1}{c}{$\leq$64k} \\
\midrule
SmolTalk2 SFT & \underline{39.82} & 23.46 & 37.93 & 36.96 \\
Plain-Paper CPT & 35.89 & 24.31 & \underline{43.10} & 38.04 \\
\textbf{Our-Data-4B CPT} & 38.45 & \textbf{25.37} & 37.93 & 35.87 \\
\textbf{Our-Data-9B CPT} & 39.96 & 24.47 & 40.52 & \underline{39.13} \\
\textbf{Our-Data-27B CPT} & \textbf{43.01} & \underline{24.69} & \textbf{43.97} & \textbf{40.22} \\
\bottomrule
\end{tabular}
\caption{Paper comprehension and long context results. Every CPT model receives SFT on the no-think subset of SmolTalk2; the first row applies the same SFT directly to the base model. All metrics are on a 0--1 scale and shown $\times$100. Best score per column in bold, second best underlined.}
\label{tab:longctx}
\end{table}

\paragraph{Trajectories improve paper comprehension.} Qasper and QASA are closely related to the synthesized corpus. On Qasper, Our-Data-27B scores over 3 points above direct SFT, while Plain-Paper CPT scores nearly 4 points below it. On QASA, all three Our-Data models outperform both direct SFT and Plain-Paper CPT.

\paragraph{Gains extend to longer contexts.} Our-Data-9B and Our-Data-27B outperform direct SFT in both LongBench v2 brackets: 9B by 2.6 and 2.2 points, and 27B by 6.0 and 3.3. Plain-Paper CPT drops by 5.06 when the bracket widens to $\leq$64k, the largest drop in the table. Greater exposure to long sequences may contribute to the trajectory models' gains: about 94\% of raw papers fit within 32K tokens, whereas about a third of the trajectories fall in the 32--64K range (\Cref{fig:length}).

\subsection{Training Dynamics}
\label{sec:dynamics}

\begin{figure}[!ht]
\centering
\includegraphics[width=0.92\textwidth]{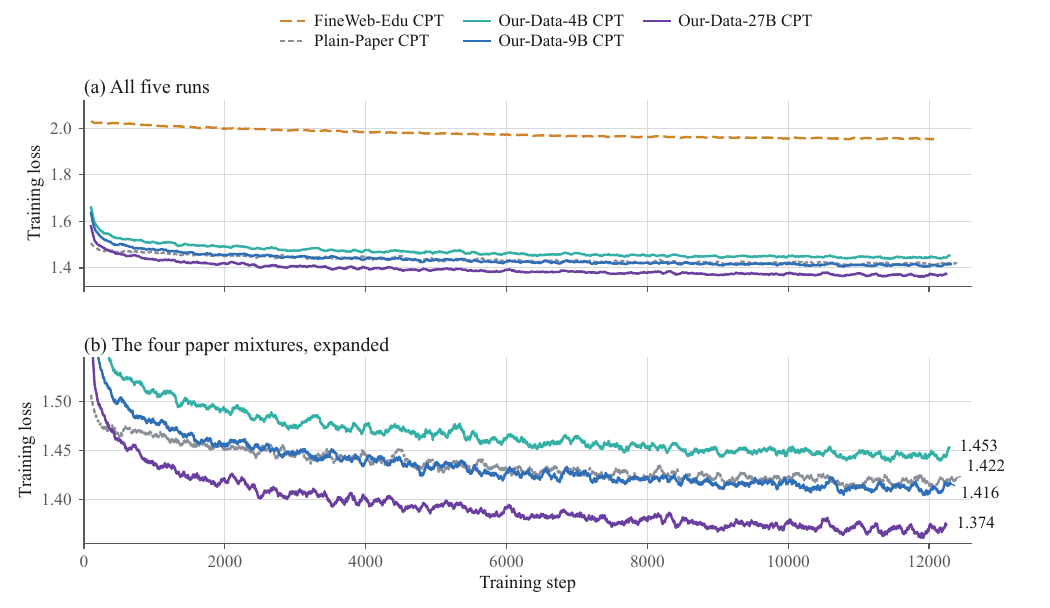}
\caption{CPT training loss, measured as cross entropy on each run's own training mixture. \textbf{Top:} All five runs. \textbf{Bottom:} The four paper mixtures with an expanded vertical scale. Curves are moving averages over 100 steps.}
\label{fig:ctloss}
\end{figure}

\paragraph{Training loss decreases with generator size.} CPT loss is 1.453, 1.416, and 1.374 for data from the 4B, 9B, and 27B generators, respectively. Each loss is measured on its own training mixture, with the mixtures differing in which generator produced their 30B trajectory tokens. The style differences in \Cref{sec:cpt} offer one possible explanation: the larger generator uses more uniform prose and less conversational deliberation, which may make its text easier to predict.

\paragraph{Higher loss accompanies stronger writing results.} The 4B corpus is the only paper mixture with higher loss than raw paper text (1.453 vs.\ 1.422). It also gives the best writing Avg.\ with DeepWriting SFT, at 54.34, and the best reasoning Avg., at 46.82. Both the 4B and 27B mixtures contain 30B trajectory tokens; the 4B mixture has higher training loss and higher writing and reasoning averages. Related work reports benefits from removing examples that are easy to predict \citep{marion2023lessismore,ankner2025perplexed}, generating reasoning data with smaller models at matched compute \citep{bansal2025smaller}, and using smaller generators for synthetic pretraining data \citep{kang2025demystifying}. Our experiments show a similar pattern across generator sizes within one construction pipeline.

\paragraph{Loss depends on the corpus and the evaluation task.} FineWeb-Edu has the highest loss, 1.953, alongside the lowest writing and reasoning averages among the CPT runs. Its broad web content also differs from the paper mixtures, illustrating how training distributions affect loss comparisons \citep{gao2021pile,muennighoff2023dataconstrained,du2024emergent}. Within the trajectory mixtures, the best generator choice depends on the task: 4B leads on writing and reasoning Avg., while 27B leads on Qasper and both LongBench v2 brackets. Together, these results favor evaluating a corpus on the intended tasks when choosing a generator.

%% file: tables/judge_robustness.tex
\begin{table}[!ht]
\centering
\footnotesize
\setlength{\tabcolsep}{5pt}
\begin{tabular}{@{}lcccc@{}}
\toprule
Model & GPT-5.5 & Gemini 3.1 Pro & Seed 2.0 Pro & Mean \\
\midrule
Qwen2.5-7B-Instruct & 50.95 & 45.18 & 59.52 & 51.88 \\
LongWriter-8B & 51.13 & 45.78 & 59.62 & 52.17 \\
DeepWriting SFT & 51.90 & 47.12 & 62.47 & 53.83 \\
FineWeb-Edu CPT + DeepWriting SFT & 51.08 & 46.00 & 60.98 & 52.68 \\
Plain-Paper CPT + DeepWriting SFT & 52.12 & 47.63 & 63.36 & 54.37 \\
\textbf{Our-Data-4B CPT} + DeepWriting SFT & \underline{54.34} & \textbf{53.49} & \textbf{67.45} & \textbf{58.43} \\
\textbf{Our-Data-9B CPT} + DeepWriting SFT & 53.48 & 50.29 & 65.11 & 56.29 \\
\textbf{Our-Data-27B CPT} + DeepWriting SFT & 53.59 & 51.30 & 65.57 & 56.82 \\
\textbf{Our-Mixed SFT} & 52.35 & 49.19 & 63.76 & 55.10 \\
\textbf{Our-Data-27B CPT} + \textbf{Our-Mixed SFT} & \textbf{54.38} & \underline{52.36} & \underline{66.41} & \underline{57.72} \\
\bottomrule
\end{tabular}
\caption{Writing Avg.\ under three judges. Mean is the equally weighted average across judges.}
\label{tab:judge-robustness}
\end{table}

%% file: sections/conclusion.tex
\FloatBarrier
\section{Conclusion and Limitations}
\label{sec:conclusion}
\enlargethispage{\baselineskip}

We presented a pipeline that unfolds scientific papers into generation trajectories with multiple turns, reconstructing a writing request, a global plan, and deliberation for each section around the paper's own text. Applied to 1.8M arXiv papers, it expands 30B tokens of paper text into 60B tokens for CPT. The same approach produces 200K SFT samples and 2,940 PAW-Bench tasks. On Qwen2.5-7B, trajectory CPT improves writing under two SFT recipes while maintaining reasoning performance close to direct SFT, with further gains in paper comprehension and long context results. Its writing gains exceed those from the same papers as plain text. All three generator sizes produce useful training data, and the 4B generator offers an effective option at lower generation cost.

Several limitations remain:
\begin{itemize}
    \item Our CPT experiments cover one base model (Qwen2.5-7B) and one mixture (30B trajectory tokens plus 20B general text tokens). Larger models and other mixture ratios remain to be evaluated.
    \item Writing Avg.\ gains hold under GPT-5.5, Gemini 3.1 Pro, and Seed 2.0 Pro. Human evaluation is still needed to assess how well these comparisons reflect reader preferences. Automated checks, sampled human inspection, and code verification of checklist items support benchmark quality.
    \item The paper comprehension and long context evaluation sets are small, so these gains provide supporting evidence alongside the writing and reasoning results.
    \item The trajectories reconstruct a plausible writing process from the finished paper; the authors' actual deliberation remains unknown.
\end{itemize}

%% file: sections/appendix.tex
\section{Scope and Overview}
\label{app:scope}

This appendix describes source processing, continued pretraining (CPT)
trajectories, SFT data construction, PAW-Bench, and the training and
evaluation protocols. Counts come from the final datasets and experiment
records, with more precise values for quantities rounded in the main text.

\paragraph{Three products from one source family.}
We produce (i) multi-turn writing trajectories for CPT,
(ii) instruction data for SFT, and (iii) an academic writing benchmark.
All three draw their factual content from scientific papers. The CPT and SFT
sources cover papers posted from 2006 through January 2026, with stricter
quality filtering for SFT. PAW-Bench uses papers posted from February through
June 2026, after the training window.

\paragraph{Terminology.}
We use \emph{CPT} for continued pretraining on autoregressive document
sequences and \emph{SFT} for supervised fine-tuning on user--assistant
examples. ``Synthetic'' refers to the reconstructed request, plan,
deliberation, or instruction. The scientific claims and prose retained as
targets originate in real papers.

\section{Paper Acquisition, Recovery, and Filtering}
\label{app:source}

\subsection{Source Recovery}

For each paper source archive we:
\begin{enumerate}
    \item locate the primary \LaTeX{} file;
    \item recursively resolve local \texttt{\textbackslash input} and
    \texttt{\textbackslash include} directives;
    \item expand locally defined macros with deterministic expansions;
    \item recover the title, abstract, ordered section hierarchy, and body text;
    \item remove content whose interpretation depends on unavailable visual or
    bibliographic context.
\end{enumerate}
The cleaning stage removes figure environments, image commands, citation
markers, layout commands, acknowledgments, references, funding statements,
and trailing template fragments. Equations and tables containing text are retained.
Appendices are excluded from the training source used in the reported
experiments. When source recovery is ambiguous, the paper is rejected.

\subsection{Deterministic Structural Filters}

The shared filtering stage removes empty or malformed bodies, errata,
supplements, duplicate bodies, unsupported languages, papers outside the
configured length interval, and papers with fewer than three substantive
sections. It also checks the lengths of the introduction and final section,
the proportions of empty sections and repeated paragraphs, residual image and
citation commands, unbalanced or incomplete \LaTeX{} environments, and unusually
frequent layout commands. Before synthesis, all training data are checked for overlap with every
downstream benchmark. For Qasper and QASA, we remove all benchmark source
papers from the CPT and SFT source pools using available paper identifiers and
normalized titles. For the remaining benchmarks, we normalize benchmark
questions and cleaned training text for case and whitespace, apply normalized
exact substring matching and token n-gram overlap, and discard the entire
training paper whenever it contains a complete benchmark question or exceeds
the overlap threshold.

The exact thresholds for SFT source papers are: abstract length at least 150 characters;
cleaned body length 8,000--180,000 characters and at least 1,000 words; at least
three logical sections; non-empty introduction; for bodies shorter than 15,000
characters, at least 100 characters in the introduction; at least 30 characters
after the final section heading; an alphabetic character ratio of at least 0.25; at
most 400 \LaTeX{} commands and 250 dollar signs per 1,000 words; duplicate
paragraph ratio at most 0.10; and a maximum paragraph repetition count of three.
The retained languages are Chinese and English. Residual acknowledgments,
references, image or citation commands, control characters, HTML fragments, and
unclosed tail environments cause rejection after cleaning.

For CPT, a small model rejects clearly defective papers,
leaving approximately 1.8M papers. SFT uses a stricter quality
assessment over the same 2006--January 2026 window. After deterministic filtering, an
LLM scores intrinsic paper quality on a 1--10 scale, emphasizing technical
soundness, novelty, clarity, organization, evidence, and writing quality.
Papers scoring below 8 are discarded, and about 356K papers meet the threshold.
The retained pool is balanced across broad arXiv subject groups by keeping
all papers from smaller groups and the highest-scoring papers from the largest
groups, producing the pool of about 218K papers in Table~\ref{app:sft-source}. Retained
bodies contain 8,000--180,000 characters, at least 1,000 words, and at least
three substantive sections.

The scoring prompt asks for one integer and a justification of two to four sentences.
Scores 9--10 are reserved for rare, particularly strong papers. The assessment
covers scientific rigor, support for the claims, novelty, coherence, fluent
writing, and the quality of the recovered text. Topic difficulty, paper length,
and standard citation practice carry no score adjustment. Balancing is deterministic after
scoring. All examples from smaller subject groups are retained. Physics and
Mathematics are capped by ranking papers by quality score and retaining the
highest-scoring portion.

\begin{table}[t]
\centering
\small
\begin{tabular}{lr}
\toprule
SFT source subject group & Papers \\
\midrule
Computer Science & 65.0K \\
Physics & 64.4K \\
Mathematics & 64.4K \\
Statistics & 11.1K \\
Electrical Engineering and Systems Science & 5.8K \\
Economics & 2.6K \\
Quantitative Biology & 2.3K \\
Quantitative Finance & 2.1K \\
\midrule
Total & $\sim$218K \\
\bottomrule
\end{tabular}
\caption{Final paper pool used as the source for SFT synthesis.}
\label{app:sft-source}
\end{table}

\section{Multi-Turn Trajectory Construction for CPT}
\label{app:cpt-pipeline}

\subsection{Trajectory Definition}

We use the trajectory $\tau(P)$ defined in \Cref{eq:trajectory}. The planning
block $p$ combines the generated global plan $g$ and section summaries used
as outlines. The original section texts $s_i$ and abstract $a$ are preserved
verbatim in the raw trajectory, before the training conversion and length
truncation described in \Cref{app:cpt-conversion}.

\subsection{Generation Stages}

The pipeline uses the following stages. The final trajectory follows paper
order; the global plan, section deliberations, and abstract deliberation can
be generated in parallel once the summaries and request are available.
\begin{enumerate}
    \item \textbf{Section summaries.} Each section is summarized independently
    to describe its role, claims, methods, and connection to the paper.
    \item \textbf{Writing request.} Given the title and all section summaries,
    the generator infers a plausible request to the author whose answer would
    be the complete paper.
    \item \textbf{Global plan.} Given the request and summary block, the
    generator reconstructs an overall strategy and an outline for each section.
    \item \textbf{Section deliberations.} Before each original section, the
    generator receives that section's real text together with summaries of the
    preceding and following retained sections. It then writes a plan for the
    section, explaining its purpose and connection to the surrounding argument.
    \item \textbf{Abstract deliberation.} The generator receives all section
    summaries and the real abstract, then explains how to compress the paper's
    problem, method, evidence, and conclusion into that abstract.
\end{enumerate}

For a paper with $n$ sections, the reported pipeline uses $2n+3$ model calls:
$n$ summaries, one request, one plan, $n$ section deliberations, and one
abstract deliberation.

\subsection{Prompt Requirements}

The prompts impose four requirements.
\begin{enumerate}
    \item The generator reasons as an author planning the paper and
    develops the section's purpose, evidence, and transitions.
    \item Every numerical result, citation, and claim must be recoverable from
    the paper context.
    \item The global plan must cover every retained section in order.
    \item A section deliberation must discuss rhetorical purpose, scientific
    content, and transition decisions in its own words without quoting the section.
\end{enumerate}
Structured section labels and summaries guide each decision and preserve
the paper's organization during trajectory construction.

\subsection{Generator Settings and Production Scale}

We produce three corpus variants with Qwen3.5-4B, Qwen3.5-9B, and
Qwen3.5-27B \citep{qwen2026qwen35}. Generation uses temperature 1.0,
top-$p$ 0.95, top-$k$ 20, presence penalty 1.5, repetition penalty 1.0,
and an 8,192-token output budget per model call. The outputs of these calls
are interleaved with the source sections into one trajectory, capped at the
128K generation limit reported in the main text. Thinking in the chat template is
disabled because the requested deliberation is itself the visible training
text.

All three generators complete trajectories for over 99.5\% of the input papers, yielding 57--60B training tokens per generator. The paired original papers contain approximately 30B tokens before unfolding.
The median source body length is 11.2K tokens, compared with trajectory
medians of 29.3K, 29.4K, and 28.3K for the 4B, 9B, and 27B generators,
respectively.
The three generators produce similar document structure and distinct
local deliberation styles.

\subsection{Training Data Preparation}
\label{app:cpt-conversion}

The cleaned title, abstract, and ordered sections from
Section~\ref{app:source} enter trajectory synthesis directly. The resulting
training sequence begins with the reconstructed writing request, which is masked
from loss, followed by the assistant's plan, section deliberations interleaved
with the corresponding original sections, and the abstract deliberation followed
by the original abstract. Documents are tokenized with the Qwen tokenizer,
truncated from the right to 65,536 tokens, shuffled, and greedily packed into 65,536-token
training sequences.

\section{SFT Data: Deriving Tasks Backwards from Paper Text}
\label{app:sft-pipeline}

\subsection{Reverse Construction}

For open-ended writing, construction begins with an answer based on the source
paper. The paper supplies the answer's factual core, and the generator applies the
minimum cleaning, reorganization, or audience adaptation needed for the task.
The request is then inferred from the resulting answer, including the length
and format constraints the answer actually satisfies. Finally, the generator
writes a deliberation with the request, completed answer, and a source
excerpt available during construction. The prompt asks for a plan based on the user's task and available material,
without references to the supplied answer. The student receives only the request
and any attached material as input; deliberation and answer form its target,
as formalized in \Cref{eq:sft-construction}.

Three distinct task types are sampled for each source paper.
Sampling weights prioritize academic writing, followed by long-form writing and
research tasks based on papers. Difficult task types that yield few examples
of sufficient quality receive additional weight.

\paragraph{Generator configuration.}
SFT construction runs Qwen3.6-27B \citep{qwen2026qwen36} at native BF16
precision. Visible thinking is enabled. Sampling uses
temperature 1.0, top-$p$ 0.95, top-$k$ 20, minimum probability 0, presence
penalty 0, repetition penalty 1.0, a 128K context window, and a 32,768-token
output budget. The SFT corpus is generated from the SFT source pool
in Table~\ref{app:sft-source}; PAW-Bench remains held out from generator
training.

\subsection{Three Construction Families}

\paragraph{A: rewriting.}
The pipeline selects a source span, transforms it into the deliverable, derives
the request from that deliverable, and then reconstructs the deliberation. These
tasks preserve most source wording and vary audience, persona, length,
organization, and output format.

\paragraph{B: paper-grounded QA and extraction.}
The pipeline jointly derives a question and its answer from a paper excerpt or
the full paper, then reconstructs the deliberation. Outputs include free-form
answers, evidence sets, records, tables, and structured extraction.

\paragraph{C: reorganization.}
The pipeline reorganizes paper content into a new form, such as
a decision memo or reading map, before deriving the request and deliberation.
These examples teach the organization of longer responses using scientific
content from the paper.

\subsection{Dual Scoring}

Each PAW-Bench task has two independent sets of grading criteria.
\begin{enumerate}
    \item \textbf{Rubric.} Four to six prompt-specific dimensions are assigned
    explicit weights. The judge scores each dimension from 1 to 10; the weighted
    mean is multiplied by 10, as in \Cref{eq:paw-scores}.
    \item \textbf{Checklist.} Three to six concrete requirements are scored as
    pass/fail. Whenever possible, deterministic checks implement word ranges,
    required headings, list counts, table dimensions, fixed endings, and other
    format constraints. Items that require an assessment of meaning use the LLM judge. The
    final checklist score is the weighted fraction passed.
\end{enumerate}
Across PAW-Bench there are 14,848 rubric criteria and 15,105 checklist
items. Of the checklist items, 8,896 (58.9\%) are deterministic and 6,209 use
the judge.

\subsection{Quality Audit}

We audited the final task pool through repeated automated checks and manual
spot checks, correcting the issues found in each round. The final audit
classified every item as answerable and scoreable, with minor issues remaining.

\section{Abbreviated Prompts, Algorithms, and Examples}
\label{app:prompts}

The templates below retain every instruction that affects the generated
target or the decision to accept it. We omit repeated formatting reminders
and long paper contexts.

\subsection{CPT Prompt Templates}

\paragraph{Section summary.}
\emph{Given the title and the current section text, write a detailed summary of
the section's key points, methods, results, and logical role in the paper.
Return only the summary.}

\paragraph{Writing request.}
\emph{Given the paper title and ordered section summaries, infer a natural
request to the author that would plausibly lead to this paper. State the
research goal, central method, and intended evidence in two or three sentences.
Vary the tone and phrasing across samples. Return only the user request.}

\paragraph{Global and local deliberation.}
\emph{Act as the paper's author. Given the request and all section summaries,
reason in first person about the scientific problem, assumptions, alternatives,
method, experiments, expected results, section order, and transitions. Then,
before each section, use its true text plus summaries of surrounding sections
to plan what that section must establish and how it should
connect to the argument. Maintain the perspective of an author planning the
paper. Return only the deliberation.}

\paragraph{Abstract deliberation.}
\emph{Given the title, all section summaries, and the true abstract, explain in
first person how to compress the problem, contribution, evidence, and conclusion
into the abstract using the paper's existing claims. Return only the
deliberation.}

\subsection{SFT Prompt Templates}

\paragraph{Answer construction.}
\emph{Use the source passage as the factual content of the requested
deliverable. Reuse its wording wherever suitable and make the minimum
reordering, trimming, connective rewriting, or audience adaptation needed. Use
claims, entities, numbers, comparisons, citations, and conclusions
supported by the source. Satisfy the sampled genre, audience, approximate
length band, and format plan. Output only the completed answer.}
For reorganization tasks, the prompt allows a new structure while requiring
the same factual support. For QA and extraction tasks, the
model jointly returns a question and answer in a parseable schema, with every
answer claim and number supported by the supplied material.

\paragraph{Reverse request.}
\emph{Use the completed answer, task family, audience, and material and format
settings to write a natural user request. The request should specify the
length and format that the answer satisfies. State the task directly.}

\paragraph{Reverse deliberation.}
\emph{Given the request, completed answer, and source excerpt for reconstruction,
write a first-person plan for answering the user's task. Frame the plan around
the request and material available to the user. Cover the necessary content
and organization, maintain a prospective viewpoint, and do not refer to the
supplied answer.}

\paragraph{Semantic judge.}
\emph{Evaluate: (1) whether the deliberation is a credible forward plan;
(2) whether the answer satisfies all explicit format and length requirements;
(3) whether the answer is clean and self-contained; (4) whether claims and numbers
are faithful under the stated material setting; (5) whether voice and audience
fit; (6) whether the request is natural and answerable; and (7) whether the
deliverable is useful. Return a 1--10 quality score and explicit problems.
References to hidden construction inputs, hidden format requirements, empty required sections,
incomplete \LaTeX{} or references, placeholders, quantitative distortion, and
deliverable mismatch are major errors.}
Examples enter the final corpus with a score of 9 or 10, an empty problem list,
and a pass on the automatic check for major errors.

\subsection{PAW-Bench Task-Generation Prompt}

We use GPT-5.5 with medium reasoning effort to generate the questions,
rubrics, and checklists.
For each assigned task type, the task writer receives the paper quality assessment,
task definition, selected material setting, and a prompt-only setting, an
excerpt, or the full cleaned paper. The abbreviated instruction is:
\emph{Write a natural academic writing request with a feasible answer length
and explicit output requirements. When material is supplied, ground every
required claim in it. For prompt-only tasks, place all scenario-specific facts
in the request. Produce four to six independent,
weighted rubric dimensions and three to six independent binary checklist items.
Make deterministic constraints exactly checkable. Before returning the item,
verify factual support, feasibility, independence of the rubric and checklist,
absence of answer leakage, and a clear pass/fail rule for each check.}

\subsection{Production Pseudocode}

\paragraph{Algorithm 1: CPT unfolding.}
\begin{enumerate}
    \item Recover and clean title, abstract, and ordered substantive sections;
    reject the paper if deterministic source checks fail.
    \item Generate one summary per section.
    \item Generate the writing request and plan for the full paper from the title and
    ordered summaries.
    \item For each section in order, generate its prospective deliberation from
    the true section and neighboring summaries, then append the original section.
    \item Generate abstract deliberation and append the original abstract.
    \item Tokenize, truncate from the right at 64K, shuffle, and pack into 64K training
    sequences.
\end{enumerate}

\paragraph{Algorithm 2: SFT reverse construction.}
\begin{enumerate}
    \item Sample three distinct task types for a quality-filtered paper using
    production weights; sample material setting, audience, length band, and
    the complexity of the formatting requirements for each task.
    \item For rewriting and reorganization, select evidence, construct the
    answer, and derive the request from it. For QA, jointly construct the
    question and answer. Generate deliberation conditioned on the request,
    completed answer, and construction excerpt. Require a plan for answering
    the task that does not refer to the supplied answer.
    \item Run the deterministic repair and validation stage. Repair
    formatting defects with unambiguous corrections; reject empty turns, broken role order,
    unclosed thinking tags, invalid schemas, unmet task requirements,
    repetition, or unsafe residual markup.
    \item Send every example that passes validation to the semantic judge. Keep examples
    with score $\geq 9$, an empty problem list, and a pass on the check for major errors.
    Save each completed result with a stable paper and task identifier so
    generation can resume from the same point after an interruption.
\end{enumerate}

\subsection{Abridged Examples}

\paragraph{CPT example (arXiv:2106.10503).}
The inferred request asks for
\emph{``a novel framework for robust Bayesian modeling of count data that
simultaneously handles zero-inflation and extreme outliers,''} centered on a
Rescaled Beta distribution, posterior robustness, Gibbs sampling, simulations,
and crime data. The global deliberation begins by identifying the joint
zero-inflation/outlier problem and comparing the intended model against Poisson
and negative binomial alternatives. The deliberation for the introduction
plans its motivation, contribution order, and transition to the
model. The paper's original sections are then placed after these generated turns, unchanged.

\paragraph{SFT example (arXiv:2012.10440, \texttt{claim\_check}).}
The user asks whether the paper \emph{actually} uses combined ALEPH and OPAL data
to determine chiral low-energy constants and operator-product-expansion
condensates, and requests a fixed verdict line plus quoted evidence. The answer
starts \emph{``Verdict: Contradicted''} and explains that the paper presents
these quantities as a future use of the combined data and confines its own
analysis to the strong coupling. The deliberation first separates claimed
current contributions from future possibilities. The semantic judge assigns
quality 10 with an empty problem list.

\paragraph{PAW-Bench example (arXiv:2605.13070).}
The task requests exactly two parts: a single \texttt{TL;DR:} sentence of at
most 30 words, followed by \texttt{Contributions:} and exactly five bullets of
12--22 words each. The requested output is plain prose and omits equations,
citations, tables, and section-by-section commentary. Its rubric weights
evidence faithfulness 0.30, contribution
selection 0.25, compression 0.20, reader utility 0.15, and terminology 0.10.
Independent checklist items directly test the two headings, sentence and bullet
counts, word ranges, and prohibited content.

\section{Training Details}
\label{app:training}

All project training runs use seed 42.

\subsection{CPT Configuration}

All reported CPT runs initialize from Qwen2.5-7B base
\citep{qwen2024qwen25}. The main experiments with synthetic data use 30B tokens from
one unfolded corpus variant and 20B tokens of FineWeb-Edu
\citep{penedo2024fineweb}, for 50B tokens total. Controls use the same budget:
plain cleaned papers plus FineWeb-Edu, and pure FineWeb-Edu.

\begin{table}[t]
\centering
\small
\begin{tabular}{ll}
\toprule
Hyperparameter & CPT value \\
\midrule
Context / packed length & 65,536 \\
Approx. tokens per update & 4M \\
Epochs & 1 \\
Peak / minimum learning rate & $2{\times}10^{-5}$ / $2{\times}10^{-6}$ \\
Schedule & cosine \\
Weight decay / gradient clipping & 0.01 / 1.0 \\
Seed & 42 \\
\bottomrule
\end{tabular}
\caption{CPT hyperparameters for the main synthetic and control runs.}
\label{app:cpt-hparams}
\end{table}

\subsection{CPT Optimization Curves}

The loss curves of the five CPT runs, their final 100-step means, and the
pairing between final training losses and downstream results are presented and
discussed in the main text (\Cref{sec:dynamics}).

\subsection{Writing SFT Configuration}

All models trained in this study follow the official DeepWriting recipe for
the writing comparison: three epochs with a constant learning rate of
$2{\times}10^{-5}$. The official DeepWriting training mixture adds 15K
OpenThoughts examples to its 20K DeepWriting examples. We use exactly the same
15K OpenThoughts examples in every corresponding row. ``Our-Mixed SFT'' replaces
10K of the 20K DeepWriting examples with 10K writing examples from
our SFT corpus and retains the other 10K DeepWriting examples and the identical
15K OpenThoughts component.

The portion drawn from our corpus prioritizes writing tasks and retains
variation in material, length, and output format. The 10K DeepWriting
half is balanced across categories and retains long-form examples.

\subsection{Writing SFT Optimization Curves}

\begin{figure*}[t]
\centering
\includegraphics[width=0.94\textwidth]{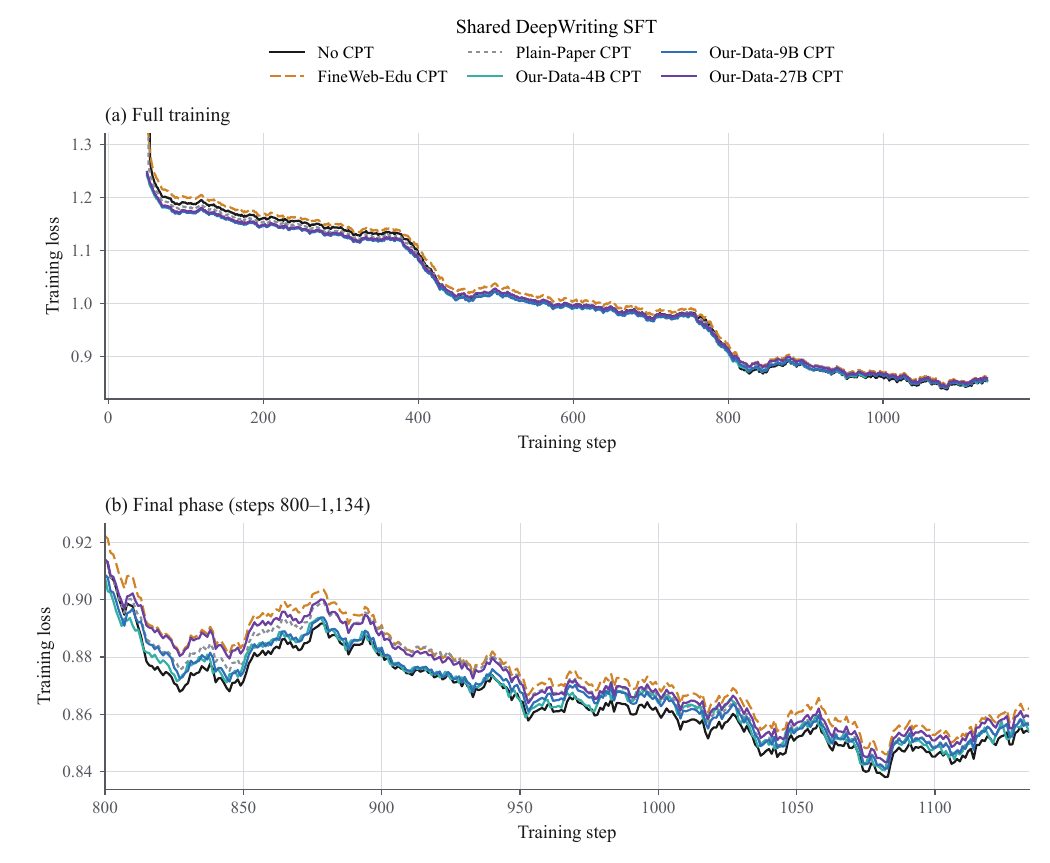}
\caption{Training loss under the shared DeepWriting SFT recipe.
\textbf{Top:} All six runs over the full training schedule.
\textbf{Bottom:} A closer view of steps 800--1,134. Curves are moving averages
over 50 steps. The runs follow similar patterns across epochs and converge
to a narrow loss range. Their downstream ranking differs from the loss ordering.}
\label{app:deepwriting-loss}
\end{figure*}

Figure~\ref{app:deepwriting-loss} compares the six runs that use the same
DeepWriting SFT mixture and recipe. All curves are 50-step moving averages over
the training loss and share the same 1,134-step axis. The lower panel enlarges
steps 800--1,134, where the curves are otherwise difficult to distinguish.

The final 50-step means are 0.854 for DeepWriting SFT, 0.862 for FineWeb-Edu
CPT + DeepWriting SFT, 0.857 for Plain-Paper CPT + DeepWriting SFT, and
0.854, 0.856, and 0.859 for the Our-Data-4B/9B/27B CPT runs, respectively.
DeepWriting SFT and Our-Data-4B CPT tie at the lowest value, while
Our-Data-4B CPT scores substantially higher on downstream writing Avg.
The SFT loss measures fit to the training mixture; the downstream benchmarks
measure how well writing skills transfer to other tasks.

FineWeb-Edu CPT, which has the highest loss throughout most of training and
the highest final mean, also has the lowest downstream Avg.\ among the runs
using DeepWriting SFT. Plain-Paper CPT lies between this control and the runs using
our CPT data. This pattern suggests that training on academic papers gives a
closer starting point for writing SFT than training on general web text.
The stronger downstream scores of our CPT runs further suggest a benefit from
reconstructing the writing process. These comparisons describe associations
between training loss and downstream performance.

\begin{figure*}[t]
\centering
\includegraphics[width=0.94\textwidth]{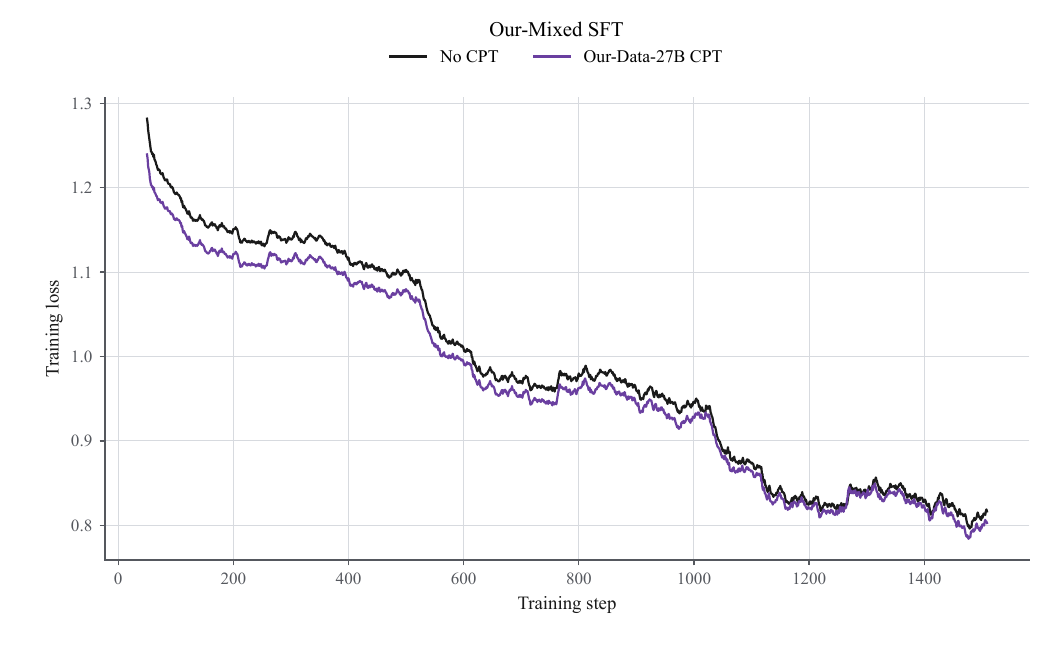}
\caption{Training loss under Our-Mixed SFT. Our-Data-27B CPT lowers the loss
across training relative to direct Our-Mixed SFT. The lower loss accompanies
a higher downstream writing Avg. Curves are 50-step
moving averages.}
\label{app:our-mixed-loss}
\end{figure*}

Figure~\ref{app:our-mixed-loss} shows the corresponding comparison under
Our-Mixed SFT. Our-Data-27B CPT remains below direct Our-Mixed SFT throughout
training and ends at 0.803, compared with 0.816. This direction agrees with the
downstream Avg., which rises from 52.35 to 54.38. Together, the lower loss
and higher score suggest that our CPT data prepares the model for this mixed
writing objective. Each figure has its own absolute loss scale because
the two SFT mixtures contain different examples and target distributions.

\subsection{General Reasoning and Paper-Reading SFT}

For the general reasoning table, all CPT checkpoints receive the same
OpenThoughts SFT \citep{guha2025openthoughts}.
For paper QA and long-context reading, all checkpoints receive the same
SmolTalk2 SFT recipe with thinking disabled. Within each table, every row uses
the same SFT data and recipe to compare the CPT initializations.

\section{Writing Benchmark Protocols}
\label{app:evaluation}

\paragraph{WritingBench.}
We use the benchmark's direct writing requests and five grading criteria
defined for each prompt \citep{wu2025writingbench}. We report the
overall score and the Academic \& Engineering subset separately.

\paragraph{PAW-Bench.}
We report both the rubric score on a 100 point scale and the 0--1 checklist pass rate on a
fixed random sample of 300 tasks from the full set of 2,940. Sampling is at
the task level, without replacement or stratification, using
\texttt{random.Random(42).sample(range(2940), 300)}; selected indices are
then sorted into their original order. All models and judges use the same
subset. It contains tasks from 248 papers: 146 tasks require no source
material, 83 provide the full paper, and 71 provide excerpts. The counts and
distributions in \Cref{tab:pawtypes,fig:pawmix} describe the full benchmark.

\paragraph{HelloBench.}
We report the Open-Ended QA and Heuristic Text Generation subsets, the two
categories closest to the paper's writing setting, using their checklist
criteria \citep{que2024hellobench}.

\paragraph{LongBench-Write.}
We preserve original task prompts and explicit length targets
\citep{bai2024longwriter}. This benchmark primarily probes sustained generation
to a requested length.

\paragraph{Judges and aggregation.}
The primary results in \Cref{tab:writing} use GPT-5.5 with medium reasoning
effort as the quality judge. We also rescore the saved responses with
Gemini 3.1 Pro and Seed 2.0 Pro (\Cref{app:additional-judges}). For each
judge, Avg.\ first averages HelloBench's two reported subsets, then takes
the mean of four equally weighted terms: WritingBench Full, PAW-Bench
rubric, the HelloBench aggregate, and LongBench-Write quality. The Mean
column in \Cref{tab:judge-robustness} averages the three judges' Avg.\ scores
equally. All averages are computed before rounding. WritingBench
Academic \& Engineering and PAW-Bench checklist pass rate are reported
separately from Avg. All three judges' tables retain the original evaluation scores for the
PAW-Bench checklist, which combines programmatic checks with LLM judgments.

\clearpage
\section{Writing Results with Additional Judges}
\label{app:additional-judges}

\Cref{tab:writing-gemini,tab:writing-seed} report the complete results for
Gemini 3.1 Pro and Seed 2.0 Pro. Both judges score the same saved model
responses used in \Cref{tab:writing}, including the fixed 300 PAW-Bench
tasks. This comparison changes the quality judge while keeping the
evaluated models and responses fixed.

\input{tables/writing_gemini}
\input{tables/writing_seed}

%% file: tables/writing_gemini.tex
\begin{table}[!ht]
\centering
\footnotesize
\setlength{\tabcolsep}{3pt}
\begin{tabular}{@{}lcccccccc@{}}
\toprule
\multirow{2}{*}{Model} & \multicolumn{2}{c}{WritingBench} & \multicolumn{2}{c}{PAW-Bench} & \multicolumn{2}{c}{HelloBench} & \multicolumn{1}{c}{LongBench-} & \multicolumn{1}{c}{\multirow{2}{*}{Avg.}} \\
\cmidrule(lr){2-3}\cmidrule(lr){4-5}\cmidrule(lr){6-7}
& \multicolumn{1}{c}{Full} & \multicolumn{1}{c}{Acad.\,\&\,Eng.} & \multicolumn{1}{c}{Rubric} & \multicolumn{1}{c}{Checklist} & \multicolumn{1}{c}{Open QA} & \multicolumn{1}{c}{Heuristic} & \multicolumn{1}{c}{Write} & \\
\midrule
Qwen2.5-7B-Instruct & 37.3 & 44.5 & 58.55 & 0.620 & 35.50 & 31.81 & 51.22 & 45.18 \\
LongWriter-8B & 37.1 & 42.6 & 54.73 & 0.524 & 34.97 & \underline{37.36} & 55.14 & 45.78 \\
DeepWriting SFT & 40.7 & 45.0 & 55.83 & 0.529 & 34.18 & 32.00 & 58.89 & 47.12 \\
FineWeb-Edu CPT + DeepWriting SFT & 39.0 & 41.9 & 51.38 & 0.535 & 35.03 & 30.78 & 60.66 & 46.00 \\
Plain-Paper CPT + DeepWriting SFT & 41.0 & 44.3 & 56.75 & 0.540 & 34.72 & 31.87 & 59.55 & 47.63 \\
\textbf{Our-Data-4B CPT} + DeepWriting SFT & \textbf{47.6} & \textbf{54.7} & 64.90 & 0.551 & \textbf{38.29} & \textbf{38.24} & \textbf{63.19} & \textbf{53.49} \\
\textbf{Our-Data-9B CPT} + DeepWriting SFT & 44.8 & 51.4 & 62.66 & 0.555 & \underline{36.30} & 31.82 & 59.58 & 50.29 \\
\textbf{Our-Data-27B CPT} + DeepWriting SFT & 43.8 & 49.5 & 63.97 & 0.557 & 35.06 & 36.24 & \underline{61.74} & 51.30 \\
\textbf{Our-Mixed SFT} & 39.4 & 45.6 & \underline{67.37} & \textbf{0.650} & 35.24 & 29.73 & 57.50 & 49.19 \\
\textbf{Our-Data-27B CPT} + \textbf{Our-Mixed SFT} & \underline{45.2} & \underline{51.7} & \textbf{69.88} & \underline{0.637} & 36.12 & 34.23 & 59.20 & \underline{52.36} \\
\bottomrule
\end{tabular}
\caption{Writing results judged by Gemini 3.1 Pro on the responses evaluated in \Cref{tab:writing}. PAW-Bench uses the same 300 tasks, and Checklist scores are retained from the original evaluation. Avg.\ follows \Cref{tab:writing}. Best scores are bold and second best underlined.}
\label{tab:writing-gemini}
\end{table}

%% file: tables/writing_seed.tex
\begin{table}[!ht]
\centering
\footnotesize
\setlength{\tabcolsep}{3pt}
\begin{tabular}{@{}lcccccccc@{}}
\toprule
\multirow{2}{*}{Model} & \multicolumn{2}{c}{WritingBench} & \multicolumn{2}{c}{PAW-Bench} & \multicolumn{2}{c}{HelloBench} & \multicolumn{1}{c}{LongBench-} & \multicolumn{1}{c}{\multirow{2}{*}{Avg.}} \\
\cmidrule(lr){2-3}\cmidrule(lr){4-5}\cmidrule(lr){6-7}
& \multicolumn{1}{c}{Full} & \multicolumn{1}{c}{Acad.\,\&\,Eng.} & \multicolumn{1}{c}{Rubric} & \multicolumn{1}{c}{Checklist} & \multicolumn{1}{c}{Open QA} & \multicolumn{1}{c}{Heuristic} & \multicolumn{1}{c}{Write} & \\
\midrule
Qwen2.5-7B-Instruct & 54.8 & 61.0 & 74.52 & 0.620 & 39.99 & 55.41 & 61.08 & 59.52 \\
LongWriter-8B & 54.2 & 58.6 & 69.33 & 0.524 & 41.92 & 57.21 & 65.35 & 59.62 \\
DeepWriting SFT & 60.3 & 64.4 & 68.99 & 0.529 & 41.42 & 54.77 & 72.47 & 62.47 \\
FineWeb-Edu CPT + DeepWriting SFT & 57.9 & 61.0 & 65.63 & 0.535 & 42.12 & 54.85 & 71.94 & 60.98 \\
Plain-Paper CPT + DeepWriting SFT & 59.9 & 63.9 & 71.88 & 0.540 & 41.53 & 55.02 & 73.40 & 63.36 \\
\textbf{Our-Data-4B CPT} + DeepWriting SFT & \textbf{65.5} & \textbf{68.8} & 75.20 & 0.551 & \textbf{43.78} & \textbf{61.74} & \textbf{76.35} & \textbf{67.45} \\
\textbf{Our-Data-9B CPT} + DeepWriting SFT & 62.7 & 66.5 & 74.38 & 0.555 & 43.04 & 56.16 & 73.72 & 65.11 \\
\textbf{Our-Data-27B CPT} + DeepWriting SFT & 62.5 & 66.5 & 75.02 & 0.557 & 42.48 & 57.87 & \underline{74.55} & 65.57 \\
\textbf{Our-Mixed SFT} & 57.9 & 62.5 & \underline{78.92} & \textbf{0.650} & 41.91 & 53.06 & 70.73 & 63.76 \\
\textbf{Our-Data-27B CPT} + \textbf{Our-Mixed SFT} & \underline{63.2} & \underline{67.7} & \textbf{79.42} & \underline{0.637} & \underline{43.43} & \underline{57.96} & 72.33 & \underline{66.41} \\
\bottomrule
\end{tabular}
\caption{Writing results judged by Seed 2.0 Pro on the responses evaluated in \Cref{tab:writing}. PAW-Bench uses the same 300 tasks, and Checklist scores are retained from the original evaluation. Avg.\ follows \Cref{tab:writing}. Best scores are bold and second best underlined.}
\label{tab:writing-seed}
\end{table}